\documentclass{article}

\PassOptionsToPackage{numbers, compress}{natbib}
\usepackage[preprint]{neurips_2024}

\usepackage[utf8]{inputenc} 
\usepackage[T1]{fontenc}    
\usepackage{hyperref}       
\usepackage{url}            
\usepackage{booktabs}       
\usepackage{amsfonts}       
\usepackage{nicefrac}       
\usepackage{microtype}      
\usepackage{xcolor}         
\usepackage{graphicx}
\usepackage{ragged2e} 
\usepackage{booktabs,makecell, multirow, tabularx, amsmath}

\newcommand{\rf}[1]{\textcolor{red}{#1}}  
\newcommand{\bd}[1]{\textcolor{blue}{#1}} 
\newcommand{\method}[1]{\textbf{#1}}
\usepackage{graphicx} 
\usepackage{adjustbox} 
\usepackage{tabularx} 

\title{EAM: Enhancing Anything with Diffusion Transformers for Blind Super-Resolution}

\author{
	Haizhen Xie\textsuperscript{*}, 
	Kunpeng Du\textsuperscript{*}, 
	Qiangyu Yan, 
	Sen Lu, 
	Jianhong Han,\\
	\textbf{Hanting Chen, 
		Hailin Hu,
		Jie Hu\textsuperscript{\dag}} \\ 
	Huawei Noah's Ark Lab\\ 
	xiehaizhen1@huawei.com, dukunpeng@huawei.com, hujie23@huawei.com
}

\begin{document}

\maketitle

\begin{abstract}
Utilizing pre-trained Text-to-Image (T2I) diffusion models to guide Blind Super-Resolution (BSR) has become a predominant approach in the field.
While T2I models have traditionally relied on U-Net architectures, recent advancements have demonstrated that Diffusion Transformers (DiT) achieve significantly higher performance in this domain.
In this work, we introduce Enhancing Anything Model (EAM), a novel BSR method that leverages DiT and outperforms previous U-Net-based approaches.
We introduce a novel block, $\Psi$-DiT, which effectively guides the DiT to enhance image restoration.
This block employs a low-resolution latent as a separable flow injection control, forming a triple-flow architecture that effectively leverages the prior knowledge embedded in the pre-trained DiT.
To fully exploit the prior guidance capabilities of T2I models and enhance their generalization in BSR, we introduce a progressive Masked Image Modeling strategy, which also reduces training costs.
Additionally, we propose a subject-aware prompt generation strategy that employs a robust multi-modal model in an in-context learning framework. This strategy automatically identifies key image areas, provides detailed descriptions, and optimizes the utilization of T2I diffusion priors.
Our experiments demonstrate that EAM achieves state-of-the-art results across multiple datasets, outperforming existing methods in both quantitative metrics and visual quality.
\end{abstract}

\section{Introduction}
\label{sec:introduction}
Blind Super-Resolution (BSR) is a fundamental and crucial task in low-level computer vision, aiming to reconstruct high-resolution (HR) images from low-resolution (LR) images without relying on prior knowledge of the image degradation process. Compared to traditional super-resolution tasks, BSR holds greater practical significance, as the degradation process of images is often unknown or complex in real-world application scenarios.
With the rapid advancement of deep learning, training effective neural networks has emerged as the dominant approach for super-resolution (SR) tasks. In the early stages, pioneering works~\cite{dong2015imagesuperresolutionusingdeep,kim2016accurateimagesuperresolutionusing,lim2017enhanceddeepresidualnetworks,Hui_2019} utilized traditional convolutional neural networks (CNNs)~\cite{oshea2015introductionconvolutionalneuralnetworks} to predict SR images. However, the quality of the generated SR images often fell short due to the limited capacity of these networks, resulting in reconstructed images that lacked fine details and exhibited noticeable artifacts.
Subsequently, Generative Adversarial Networks (GANs)~\cite{goodfellow2014generative} were introduced to address these limitations. Notably, classical SR networks such as~\cite{ledig2017photorealisticsingleimagesuperresolution, zhang2021blind, wang2021real} leveraged the adversarial training paradigm of GANs to enhance the realism and quality of generated SR images. Despite these improvements, the inherent limitations of GAN-based methods, such as mode collapse and training instability, still posed challenges in consistently producing high-fidelity SR images with rich details.

In recent years, generative modeling has witnessed remarkable progress, especially in the realm of text-to-image generation. Among these innovations, diffusion models have garnered significant attention for their exceptional ability to generate high-quality, photorealistic images.
A notable milestone in this development is Stable Diffusion and its successors, which have significantly advanced text-to-image synthesis~\cite{rombach2022highresolutionimagesynthesislatent, podell2023sdxlimprovinglatentdiffusion}.These models are primarily based on the U-Net architecture~\cite{ronneberger2015unetconvolutionalnetworksbiomedical}, a widely-used framework for tasks such as segmentation and denoising.
Moreover, there has been a growing trend toward integrating Diffusion Transformers (DiTs) into these models~\cite{peebles2023scalablediffusionmodelstransformers}. Prominent examples include~\cite{esser2024scalingrectifiedflowtransformers,li2024hunyuanditpowerfulmultiresolutiondiffusion}. 
These models leverage attention mechanisms to handle long-range dependencies and complex data relationships, thereby enhancing the generative capabilities of diffusion models. The adoption of the DiTs architecture has unveiled extraordinary potential, continually pushing the boundaries of text-to-image generation.

Driven by the remarkable achievements of diffusion models, diffusion-based methods have been increasingly adopted in the super-resolution (SR) field, with notable examples including~\cite{wang2023exploiting,lin2023diffbir, yang2023pixel, wu2024seesrsemanticsawarerealworldimage, yu2024scalingexcellencepracticingmodel}. These methods commonly adhere to the ControlNet paradigm~\cite{zhang2023addingconditionalcontroltexttoimage}, a controllable diffusion model that introduces a separate branch alongside the basic diffusion model. This design enables customization of target high-resolution (HR) images by leveraging processed low-resolution (LR) latent representations and associated semantic prompts. 
However, our experiments reveal that this paradigm limits the generative capacity of the base model, as the control branch is isolated from the main branch. Given the emergence of Diffusion Transformers (DiTs) as potent diffusion networks, developing a DiT-based SR model is essential. 
Predictably, the limitations of ControlNet will become a bottleneck for achieving excellent SR performance in DiT-based SR networks.
To address this situation, we propose Enhancing Anything Model (EAM), a unique DiT-based architecture for blind image SR. Based on a pre-trained text-to-image DiT, we construct a triple-flow DiT block, termed $\Psi$-DiT, in which the noisy latents interact with both text embeddings and LR latents through attention operations. Our ablation experiments show that the modified blocks outperform traditional ControlNet. Additionally, we employ a new training strategy to enhance the generative power and robustness of the entire model. For semantic prompts, we adopt the state-of-the-art vision-language model (VLM) Mini-CPM~\cite{hu2024minicpmunveilingpotentialsmall} with a context learning approach to provide semantic information. Extensive experiments demonstrate that our model achieves excellent pixel-level SR performance and outperforms existing SR methods on realistic test datasets.
Our main contributions can be summarized as follows
\begin{itemize}
	\item We introduce EAM, a novel framework that applies DiT models to image super-resolution tasks. This framework features a triple-flow DiT architecture that integrates the control branch, thereby enhancing the model's generative capacity and improving output expressiveness and fidelity.
	\item We propose a Progressive MIM strategy in the control branch of the triple-flow block to maximize the generative potential and robustness of diffusion models. Additionally, we enhance subject-aware prompt generation through multi-modal modeling and in-context learning.
	\item We demonstrate that EAM achieves superior perceptual quality and improved generation effects for blind image super-resolution, as evidenced by both quantitative and qualitative evaluations. This shows that EAM outperforms existing methods in terms of perceptual quality and robustness.
\end{itemize}

\section{Related Work}
\label{sec:relatedwork}
\paragraph{Traditional SR methods.}
The field of image Super-Resolution (SR) has witnessed rapid advancements in recent years, transitioning from traditional interpolation techniques to deep learning-based approaches that achieve state-of-the-art performance. SRCNN~\cite{dong2015imagesuperresolutionusingdeep} pioneered the use of Convolutional Neural Networks (CNNs) for SR tasks. Subsequent works, such as~\cite{kim2016accurateimagesuperresolutionusing,lim2017enhanceddeepresidualnetworks}, further enhanced SR performance by employing deeper networks and more sophisticated training strategies. However, these methods often produce images with limited detail, as they are trained under simplified degradation assumptions.
To address more complex SR scenarios, GANs have emerged as a powerful solution. SRGAN~\cite{ledig2017photorealisticsingleimagesuperresolution} was the first major GAN-based SR model, introducing an adversarial loss alongside a perceptual loss derived from a pre-trained deep network. This approach significantly improved the realism and quality of generated SR images. Subsequent models, such as BSRGAN~\cite{zhang2021blind} and Real-ESRGAN~\cite{wang2021real}, further advanced the field by focusing on blind SR and handling real-world images degraded by complex factors.

\paragraph{Text-to-Image Diffusion methods.}
Diffusion models have garnered significant attention in the field of generative modeling due to their ability to generate high-quality, photorealistic images. Initially, models like DDPM~\cite{ho2020denoising} demonstrated the potential of diffusion models in probabilistic modeling by simulating the process of adding noise to data and then learning to reverse this process. More recently, Stable Diffusion and its variants~\cite{rombach2022highresolutionimagesynthesislatent, podell2023sdxlimprovinglatentdiffusion} have achieved remarkable results, outperforming traditional models with their stability and capacity for detailed, realistic outputs. The introduction of Diffusion Transformers (DiT)~\cite{peebles2023scalablediffusionmodelstransformers} has further enhanced the generative capabilities of diffusion models, surpassing the performance of UNet-based architectures. DiT has been successfully applied in various tasks, such as~\cite{esser2024scalingrectifiedflowtransformers, li2024hunyuanditpowerfulmultiresolutiondiffusion}.
ControlNet~\cite{zhang2023addingconditionalcontroltexttoimage} and its variants, including IP-Adapter~\cite{ye2023ipadaptertextcompatibleimage}, ControlNet++~\cite{li2024controlnetimprovingconditionalcontrols}, and ControlNeXt~\cite{peng2024controlnextpowerfulefficientcontrol}, have provided flexible and controllable diffusion architectures for text-to-image generation. These methods leverage a replicated portion of the pre-trained diffusion model to enable more customized and context-aware image synthesis.

\paragraph{Diffusion-based SR methods.}
The impressive performance of text-to-image diffusion models has spurred the development of diffusion-based SR methods. Notable examples include~\cite{wang2023exploiting,lin2023diffbir, yang2023pixel, wu2024seesrsemanticsawarerealworldimage, yu2024scalingexcellencepracticingmodel}. These methods typically employ a ControlNet-based paradigm, which introduces an additional branch to the base diffusion model to inject processed low-resolution (LR) latents, semantic prompts, and other useful information for SR. For instance, PASD~\cite{yang2023pixel} uses LR images processed by SwinIR~\cite{liang2021swinirimagerestorationusing} as inputs to the ControlNet and injects captions or detection results into the main branch. SeeSR~\cite{wu2024seesrsemanticsawarerealworldimage} employs DAPE to extract representation embeddings and tags from LR images, which are then injected into the ControlNet and text encoder. SUPIR~\cite{yu2024scalingexcellencepracticingmodel} scales up the base model and preprocesses LR images with a degradation-robust encoder for the ControlNet, enhancing overall SR performance.
While these methods focus on enhancing the control branch, they often overlook the generative potential of the main branch. With the advent of DiT's superior text-to-image capabilities, the feasibility of the ControlNet paradigm for DiT-based SR modeling warrants reevaluation.

\section{Enhancing Anything Model}
\label{sec:method}
\begin{figure}
	\centering
	\includegraphics[width=1\linewidth]{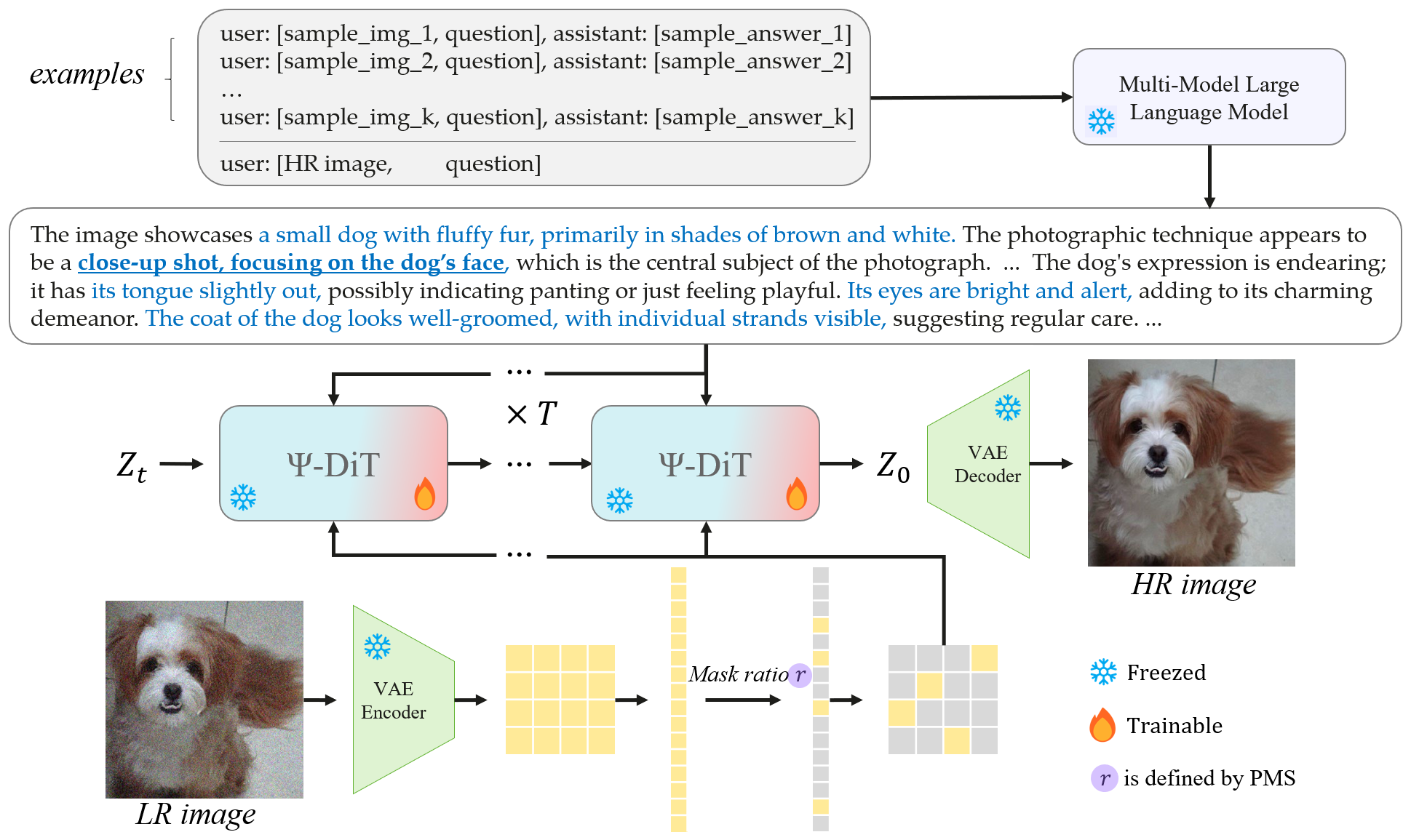}
	\caption{This figure briefly demonstrates the training workflow of our proposed EAM method.}
	\label{fig:framework}
\end{figure}

As illustrated in Fig.~\ref{fig:framework}, our proposed Enhancing Anything Model (EAM) is composed of three key components: a novel DiT-based architecture named $\Psi$-DiT, a tailored Progressive Masked Image Modeling (MIM) strategy for training, and an innovative pipeline for generating optimized prompts.
\subsection{Overall Structure}
\begin{figure}
	\centering
	\includegraphics[width=0.6\linewidth]{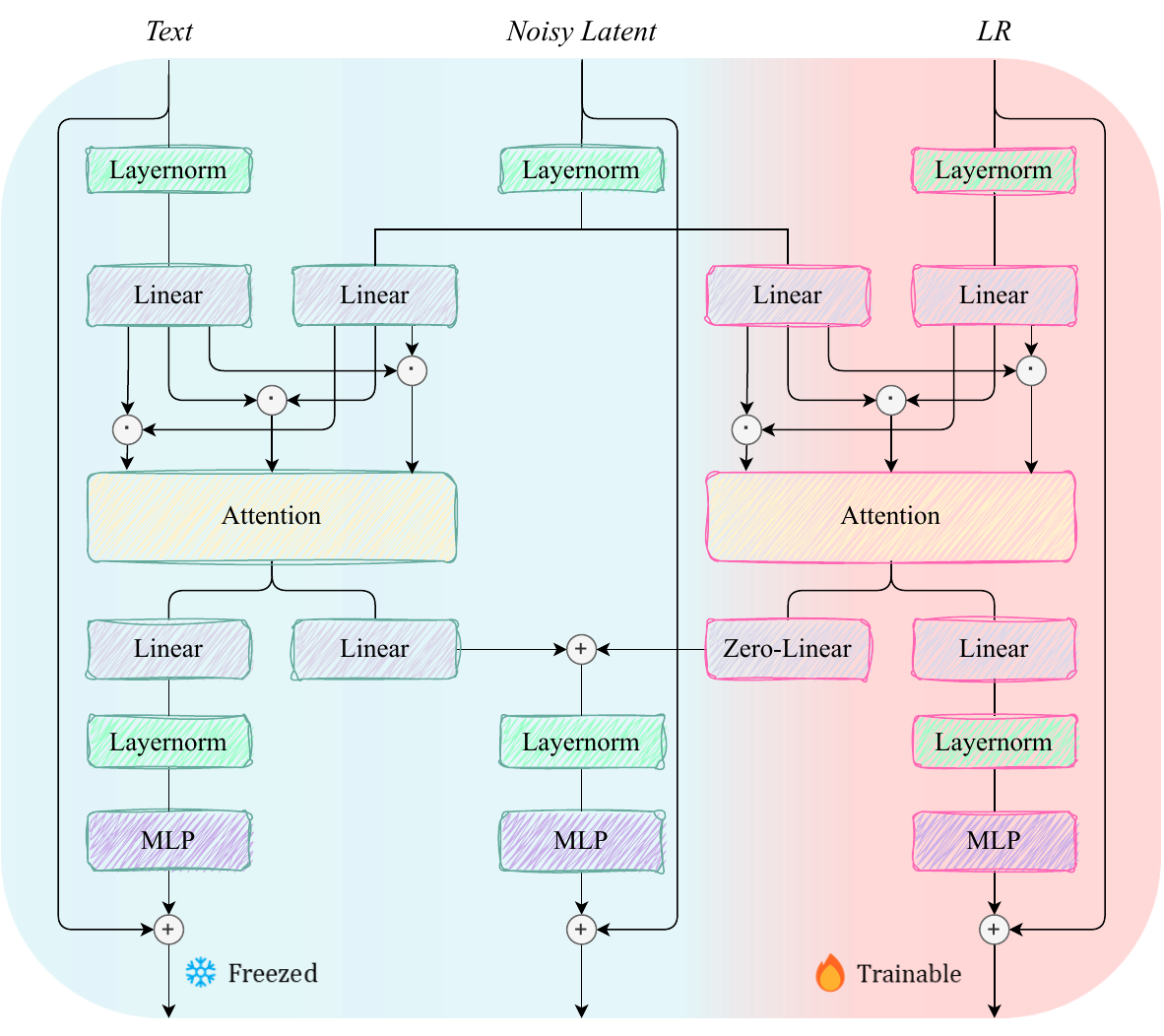}
	\caption{Detail of $\Psi$-DIT. On the left, the pre-trained MMDIT architecture remains frozen to preserve its learned prior. On the right, the trainable Separable Stream Control Module (SSCM) been designed to enhance the model's adaptability and performance.}
	\label{fig:content}
\end{figure}

\paragraph{Triple-flow Architecture $\Psi$-DiT.}
The ControlNet module~\cite{zhang2023addingconditionalcontroltexttoimage}, initially introduced by DiffBIR~\cite{lin2023diffbir}, has gained widespread adoption in real-world super-resolution applications following enhancements by PASD~\cite{yang2023pixel}, SeeSR~\cite{wu2024seesrsemanticsawarerealworldimage}, and SUPIR~\cite{yu2024scalingexcellencepracticingmodel}. It has also been adapted for use within the DiT series of models. However, we contend that the ControlNet architecture exhibits notable limitations when utilized for real-world super-resolution tasks within the DiT series~\cite{esser2024scalingrectifiedflowtransformers, li2024hunyuanditpowerfulmultiresolutiondiffusion}. Firstly, it demands the replication of the entire base model's branches, incurring substantial computational costs during training. Secondly, the direct summation of noisy latent and low-resolution (LR) latent vectors, which represent distinct modalities, constrains the model's generative potential. To mitigate these limitations, we propose a novel architecture, $\Psi$-DiT. This architecture distinctly partitions the input modalities: text tokens, noisy latent tokens, and LR tokens. Each modality is processed by dedicated DiT modules, with cross-modal interactions mediated exclusively through attention mechanisms.
As depicted in Fig.~\ref{fig:content}, the original dual-stream MMDIT architecture, driven solely by text, has evolved into a triple-flow architecture. This new design is now driven by both text and LR images. Furthermore, to fully harness the generative capabilities of the pre-trained text-to-image (T2I) model, we elect to freeze the weights of the text control branch inherited from the pre-trained T2I base model during training. This strategic choice allows us to focus on the newly integrated LR latent control branch. Ultimately, this approach confers two advantages. Firstly, compared to the full copy strategy employed by ControlNet, our $\Psi$-DiT architecture introduces only one additional branch, thereby minimizing the increase in new training parameters. Secondly, the LR latent control branch, which exerts conditional control exclusively through attention modules, is poised to maximize the utilization of the model's prior knowledge and mitigate the risk of degenerate overfitting.
\paragraph{Separate Stream Control Module.}
In the original MMDIT module, the noisy latent and text embeddings each utilize distinct linear and multilayer perceptron (MLP) layers. During the attention computation phase, the query (Q), key (K), and value (V) vectors of the noisy latent and text embeddings are concatenated along the token dimension prior to performing the attention calculation. Following output splitting, each portion is directed to its respective MLP module via a linear layer.
As illustrated in Fig.~\ref{fig:content}, we have introduced a novel stream-based separate conditional control module (SSCM), building upon the MMDIT framework. This SSCM module takes as input the tokens from both the noisy latent branch and the low-resolution (LR) latent branch. Similar to the MMDIT module, the QKV vectors of the noisy latent and LR latent are concatenated along the token dimension for attention calculation. Post-output splitting, the portion corresponding to the noisy latent is merged with the input at the noisy latent position of the MMDIT module through a linear layer. Concurrently, the output at the LR latent position is fed into the MLP module via a linear layer, serving as the input for subsequent processing blocks.
\paragraph{Weight Initialization Policy.}
In downstream tasks, weight initialization is a critical factor that influences both the model's convergence rate and its eventual generalization capabilities.
Acknowledging this, we have devised specific weight initialization strategies aimed at enhancing the model's performance and efficiency. A key factor in the remarkable success of ControlNet was the incorporation of zero-conv, which notably accelerated convergence. Drawing on this insight, we have integrated a zero-initialized layer in the final attention layer of our model. This strategic addition ensures a seamless transition from text-to-image tasks to image-to-image tasks, building on the successful precedent set by zero-conv.
Naturally, we inherit the weights of the pre-trained model for the QKV of the noisy latent, aligning with our approach. However, for the weight initialization of the image control branch, we face a choice between two strategies. The first is to inherit the weights from the text control branch, which, like the image control branch, is a conditional branch. The second option is to adopt the weights from the noisy latent branch, which shares a similar modality.
After careful deliberation, we have opted to initialize the weights of the LR latent branch by copying those of the noisy latent branch. This decision is based on the closer modality relationship between the LR latent and the noisy latent, making the transfer of weights more logical and beneficial for the model's performance.

\subsection{Progressive MIM Strategy}
Masked Image Modeling (MIM) has been introduced to many high-level vision tasks, enabling models to acquire a generic representation of images and thereby enhancing their generalization capabilities across a variety of tasks. Masked Autoencoder (MAE)~\cite{he2021maskedautoencodersscalablevision} is a typical MIM approach, capable of reconstructing the entire image from masked, partial inputs. By leveraging MIM, models can automatically infer and supplement missing information from even minimal input data. Building on this insight, we integrate MIM into our diffusion-based super-resolution network to enhance its generative capabilities.
However, the primary purpose of MIM in high-level vision tasks diverges from that in low-level vision tasks, making the direct application of MIM to low-level tasks sub-optimal. In low-level vision, the typical goal is to enhance a low-quality image to achieve high-quality output. In contrast, MIM aims to generate a complete image from a masked input. The degradation domains of MIM tasks are congruent, meaning that both the masked and restored images share similar levels of degradation or cleanliness. Consequently, when adapting MIM to low-level visual tasks such as super-resolution, it is crucial to consider how to make the network aware of the degradation process. This definitely raises the difficulties of the learning task for the network. It is challenging for the network to balance its generative capabilities and the preservation of the original image information, and there is a significant risk that this imbalance could lead to a collapse.
Inspired by these insights, we have designed a novel \textbf{P}rogressive \textbf{M}asked Image Modeling \textbf{S}trategy (PMS) that leverages curriculum learning~\cite{wang2021surveycurriculumlearning} principles. This strategy enables the model to learn masked image super-resolution from more familiar to unfamiliar examples by controlling the mask ratio. It smoothly transitions the network from a text-to-image (T2I) model to an image-to-image (I2I) model. We observed that PMS significantly enhances training stability and results in the generation of more realistic and detailed image features.
Specifically, given the mask ratio threshold  $r_{min}$, the total number of MIM training steps $t$, the number of progressive steps $k$, and the progressive clip $c$, we adjust the mask ratio based on the number of training steps $p$. When the number of training steps $p$ has not reached $k$, the mask ratio is randomly picked from a corresponding range. Once the training steps $p$ exceed $k$, the mask ratio is fixed to $r_{max}$ until the end of MIM training.
\begin{equation}
	\begin{cases}
		r = 1- ( 1 - r_{min}) \cdot (\lfloor \frac{p \cdot c}{k} \rfloor + \sigma) / c , & \text{if } p < k
		\\
		r = r_{min}, & \text{if } p \geq k
	\end{cases}
\end{equation}
Here, $\sigma \sim \mathcal N(0,1)$ is a random item. 
We use $t/2$ as a default value of $k$, and $s$  as a default value of $c$.

\begin{figure}
	\centering
	\includegraphics[width=0.6\linewidth]{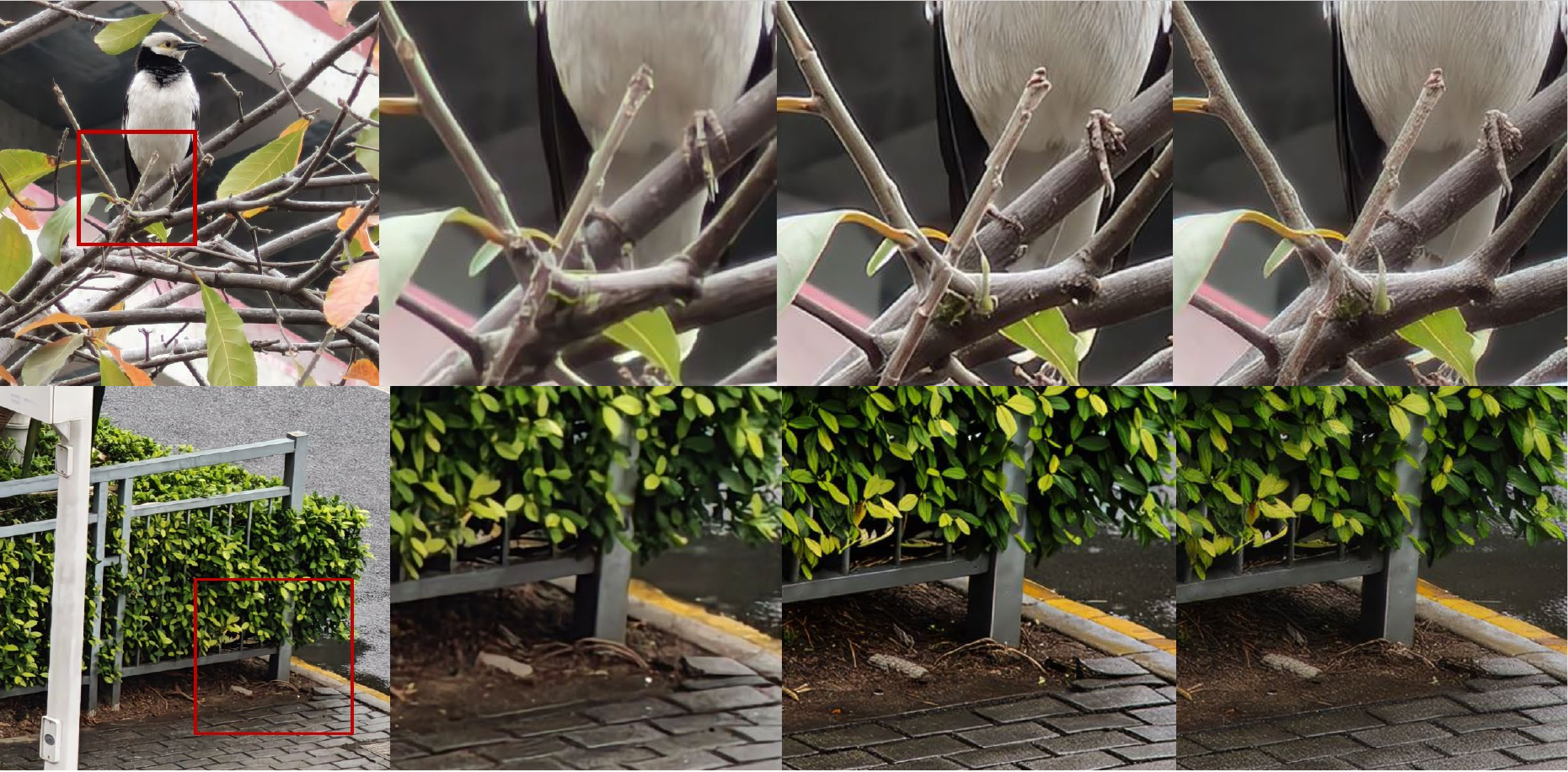}
	\caption{Restoration results of model using PMS or not. from the left to the right: the whole LR input, the cropping input region, model results trained without PMS, model results trained with PMS.}
	\label{fig:ab2}
\end{figure}

Compared with selecting random mask ratios throughout the training process, models using PMS are more likely to encounter image pairs with larger mask ratios at the beginning, aligning closely with the original T2I training paradigm. As training progresses, the model gradually adapts to conditional injections, and it becomes more likely to encounter image pairs with a lower mask ratio. This allows for a deeper exploration of the model's generative capabilities based on reference images.
In addition, we use unmasked data pairs to perform SFT, which further strengthens the model's ability to remove degradation based on its generative capabilities. Experiments have proven that using unmasked data to perform SFT can effectively improve pixel-level alignment capabilities. For specific experimental results, please refer to Tab.~\ref{tab:prompts_comp} and Fig.~\ref{fig:ab2}.

\subsection{Subject-aware Prompts Generation}
When we see an image, our brain allocates different attention to different regions within the image. 
Typically, for images with a clear focal point, the brain tends to concentrate more on the content within the focal area 
before shifting attention to the non-focal regions. 
Conversely, for images with ambiguous or no focal points, such as wide-angle shots, 
the brain tends to distribute its attention more evenly across the entire image, focusing on the overall content and detail.

Building upon this observation, we have developed a description generation pipeline that 
leverages a state-of-the-art Vision-and-Language Model (VLM) to generate prompts for training. 
In this pipeline, the VLM first identifies the focal regions of an image. 
It then proceeds to describe these focal areas with as much detail as possible. 
For the non-focal regions, the VLM generates descriptions speculatively, based on the available visual cues. 
In cases of images with no clear focus, such as wide-angle shots, 
the VLM enhances its speculative descriptions to cover the entire image, ensuring a comprehensive narrative.

As show in Fig.~\ref{fig:framework}, we have select a collection of sample images that encompass both focused and non-focused areas. 
Utilizing in-context learning, we providing a set of ideal output results at beginning. 
These samples, along with their corresponding ideal results, serve as the example inputs for our VLM model.
Then the VLM can annotate the next input image with previous given examples.
The prompts generated through this pipeline are more sensitive to the focal points of the images compared to simply describe them. 
This approach yields richer details and incorporates more speculative content, which significantly enhances the model's generation capabilities. 
For further experimental validation and results, please refer next section.

\section{Experiments}
\label{sec:Experiments}
\subsection{Experimental Settings}
\paragraph{Training Datasets.}
The training process of our Enhancing Anything Model (EAM) leverages a diverse range of datasets, including LSDIR~\cite{li2023lsdir}, DIV2K~\cite{timofte2017div2k}, Flickr2K~\cite{agustsson2017flicker2k}, OST~\cite{wang2018ost}, and the initial 10K face images from FFHQ~\cite{ffhq}.
To create training pairs of low-resolution (LR) and high-resolution (HR) images, we utilized the degradation pipline from Real-ESRGAN~\cite{wang2021real}, which effectively simulates real-world image degradation.

\paragraph{Test Datasets.}
To comprehensively evaluate theperformance of our model against various real-world super-resolution methods, we empoyed both synthetic and real-world datasetsas test sets.
Specifically:
1) We selected 100 images from the validation set of DIV2K~\cite{timofte2017div2k}, resized the short edge to 1024 pixels, and generated LR-HR pairs through a degradation pipeline consistent with the training process. This datasetis referred to as DIV2K-VAL.
2) We utilized the real-world dataset realphoto60, collected in SUPIR~\cite{Fanghua2024supir}, resizing all images to 1024 pixels for inference.
3) In line with the protocols of the latest blindimage super-resolution competition, we adopted the real-world test set from the final stage of the NTIRE 2024 RAM competition~\cite{RAIM} as our benchmark. This set comprises 50 real-life images captured on actual phones, providing a practical and accurate assessmentof our model's capabilities.

\paragraph{Implementation Details.}
Implementation Details. Our EAM is based on the SD3-medium~\cite{esser2024scalingrectifiedflowtransformers} pre-trained text-to-image model.
Training was performed using the Adam optimizer over 30,000 iterations. The entire experiment was performed with 1024-pixel resolution images and implemented on 32 NPUs.

\subsection{Comparison with Existing Methods}
\paragraph{Quantitative Comparisons.}
\begin{table}
	\begin{center}
		\scalebox{0.85}{
			\begin{tabular}{c|c|cccccc}
				\hline
				Datasets & Metrics & \makecell[c]{StableSR} & \makecell[c]{DiffBIR} & \makecell[c]{PASD} & \makecell[c]{SeeSR} & \makecell[c]{SUPIR} & \makecell[c]{EAM} \\
				\hline
				\multirow{6}{*}{\makecell[c]{DIV2K-Val}} 
				& PSNR $\uparrow$ & \rf{21.07} & 20.94 & 20.95 & \bd{21.02} & 19.93 & 19.87 \\
				& SSIM $\uparrow$ & \rf{0.5687} & 0.5398 & 0.5626 & \bd{0.5662} & 0.5189 & 0.5225  \\
				& LPIPS $\downarrow$ & 0.3767 & 0.3728 & 0.3821 & \bd{0.3321} & 0.3655 & \rf{0.3318} \\
				& ManIQA $\uparrow$ & 0.5396 & \bd{0.6085} & 0.5511 & 0.6215 & \bd{0.6321} & \rf{0.6601} \\
				& CLIP-IQA $\uparrow$ & 0.5357 & \rf{0.7258} & 0.6158 & 0.7092 & 0.6252 & \bd{0.7150} \\
				& MUSIQ $\uparrow$ & 61.20 & 70.89 & 69.12 & \bd{72.32} & 70.88 & \rf{73.95} \\
				\hline
				\multirow{3}{*}{\makecell[c]{RealWorld60}}
				& ManIQA $\uparrow$ & 0.5003 & \bd{0.6087} & 0.5232 & 0.6067 & 0.6044 & \rf{0.6333} \\
				& CLIP-IQA $\uparrow$ & 0.5278 & \rf{0.7729} & 0.6143 & \bd{0.7379} & 0.6561 & 0.715 \\
				& MUSIQ $\uparrow$ & 54.59 & \bd{69.84} & 63.14 & \bd{71.46} & 70.47 & \rf{73.95} \\
				\hline
				\multirow{3}{*}{\makecell[c]{NTIRE2024-RAM50}}
				& ManIQA $\uparrow$ & \bd{0.6986} & 0.6482 & 0.6497 & 0.6639 & 0.6531 & \rf{0.7135} \\
				& CLIP-IQA $\uparrow$ & 0.6857 & \rf{0.7034} & 0.6644 & 0.6886 & 0.5199 & \bd{0.7027} \\
				& MUSIQ $\uparrow$ & \bd{72.98} & \bd{71.87} & 72.14 & 72.79 & 67.36 & \rf{73.56} \\
				\hline
			\end{tabular}
		}
		\caption{Quantitative comparison with state-of-the-art methods on both synthetic and real-world benchmarks. \rf{Red} and \bd{blue} colors represent the best and second best performance, respectively.}
		\label{tab:metric_tab}
	\end{center}
\end{table}
As illustrated in Tab.~\ref{tab:metric_tab}, our methodology effectively addresses a wide range of image degradations. In our quantitative evaluation, we employ Real-ESRGAN's degradation protocols to rigorously benchmark our approach against state-of-the-art algorithms that share similar objectives, including StableSR~\cite{wang2023exploiting}, DiffBIR~\cite{lin2023diffbir}, PASD~\cite{yang2023pixel}, SeeSR~\cite{wu2024seesrsemanticsawarerealworldimage}, and SUPIR~\cite{yu2024scalingexcellencepracticingmodel}.

Our evaluation encompasses three distinct datasets: DIV2K, RealWorld60, and RAM50. RealWorld60 is sourced from the SUPIR dataset~\cite{Fanghua2024supir}, while RAM50 is drawn from the NTIRE2024 challenge~\cite{RAIM}. To streamline the metric computation process, we extract a central square crop from the output images. The evaluation metrics are a comprehensive set of full-reference metrics, comprising PSNR, SSIM, and LPIPS, alongside no-reference metrics such as ManIQA~\cite{yang2022maniqamultidimensionattentionnetwork}, ClipIQA~\cite{wang2022exploringclipassessinglook}, and MUSIQ~\cite{ke2021musiqmultiscaleimagequality}. Owing to the unavailability of ground truth images in RealWorld60 and RAM50, our comparative analysis relies exclusively on no-reference metrics. 

Our method exhibits satisfactory performance in full-reference metrics on the DIV2K dataset, suggesting a considerable level of fidelity. Additionally, it shows strong competitiveness in no-reference metrics across all three datasets, with only a few metrics slightly underperforming compared to DiffBIR.

\paragraph{Qualitative Comparisons.} 

\begin{figure}
	\scriptsize
	\centering
	\vspace{-4mm}
	\resizebox{\textwidth}{!}{
		\begin{tabular}{l}
			\begin{adjustbox}{valign=t}
				\begin{tabular}{c}
					\includegraphics[height=0.36\textwidth]{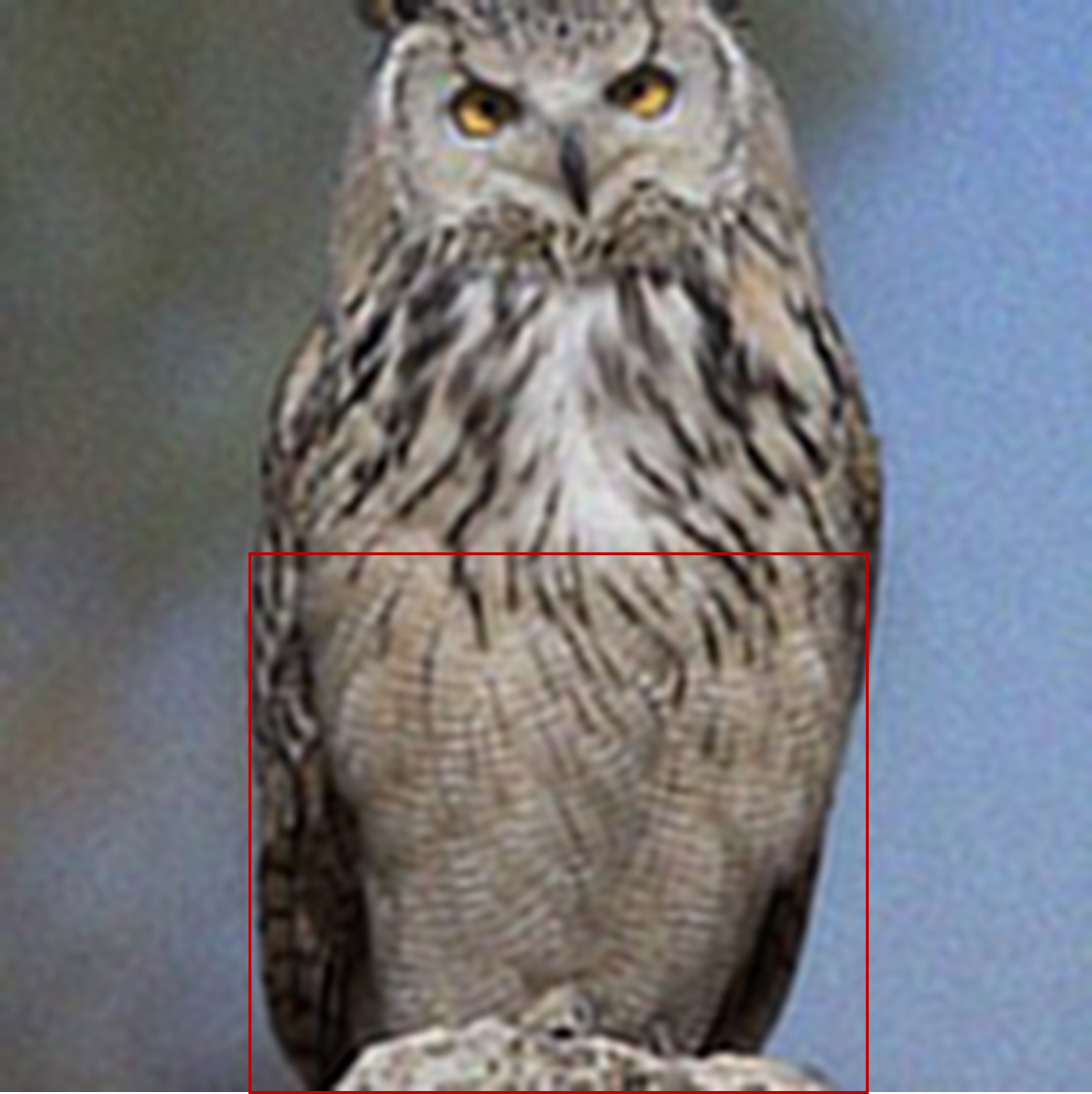} \\
					\method{LR}
				\end{tabular}
			\end{adjustbox}
			\hspace{-2mm}
			\begin{adjustbox}{valign=t}
				\begin{tabular}{ccc}
					\includegraphics[width=0.18\textwidth]{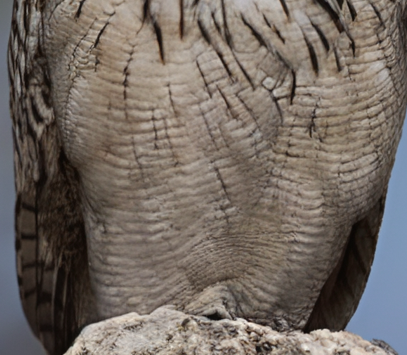} &
					\includegraphics[width=0.18\textwidth]{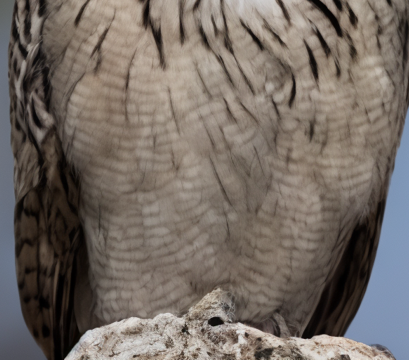} &
					\includegraphics[width=0.18\textwidth]{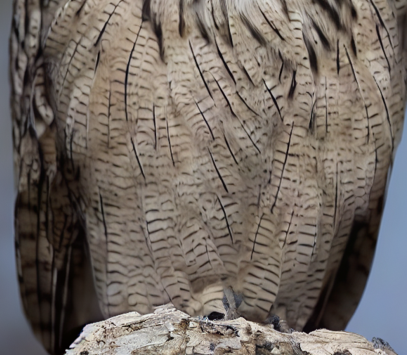} \\
					StableSR\cite{wang2023exploiting} &
					DiffBIR\cite{lin2023diffbir} &
					PASD\cite{yang2023pixel} \\
					\includegraphics[width=0.18\textwidth]{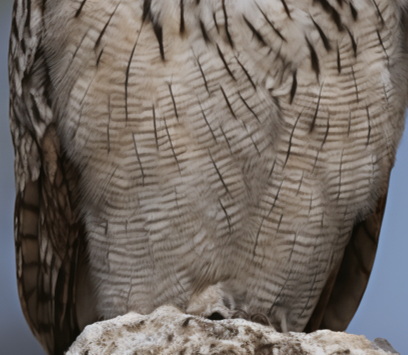} &
					\includegraphics[width=0.18\textwidth]{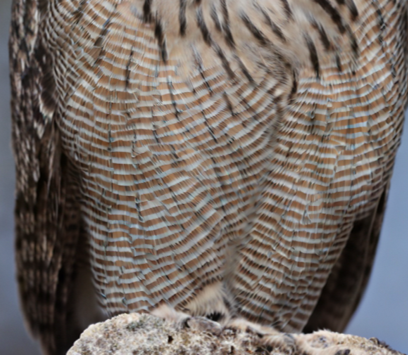} &
					\includegraphics[width=0.18\textwidth]{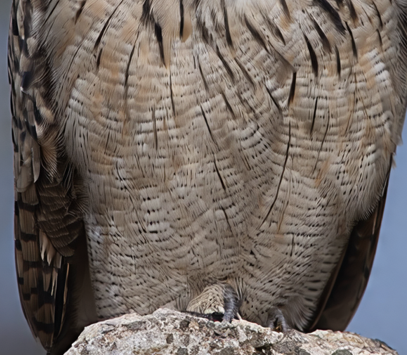} \\
					SeeSR &
					SUPIR &
					\method{EAM (ours)}
				\end{tabular}
			\end{adjustbox}
			\hspace{-2mm}
			\\
			\begin{adjustbox}{valign=t}
				\begin{tabular}{c}
					\includegraphics[height=0.36\textwidth]{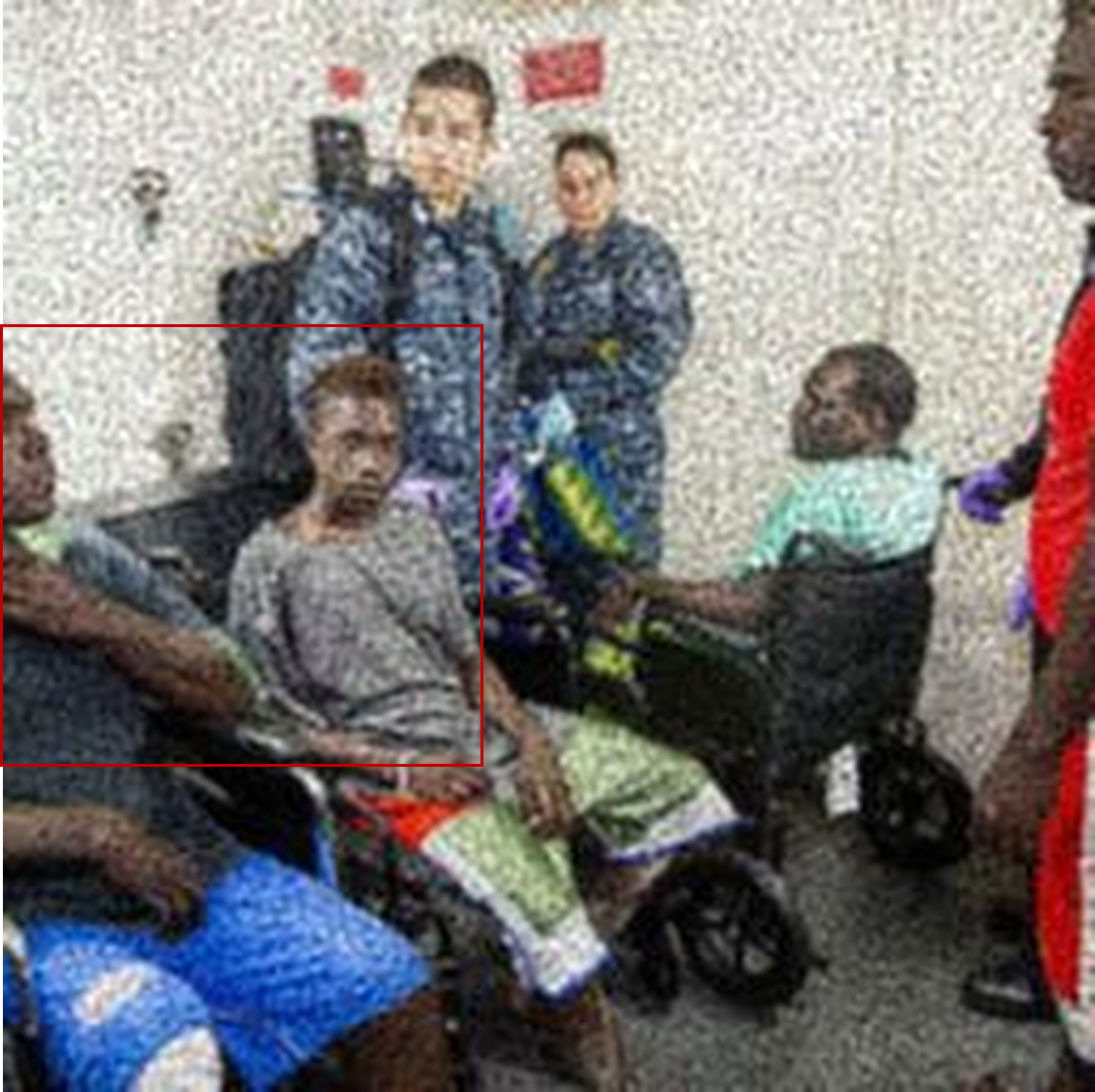} \\
					\method{LR}
				\end{tabular}
			\end{adjustbox}
			\hspace{-2mm}
			\begin{adjustbox}{valign=t}
				\begin{tabular}{ccc}
					\includegraphics[width=0.18\textwidth]{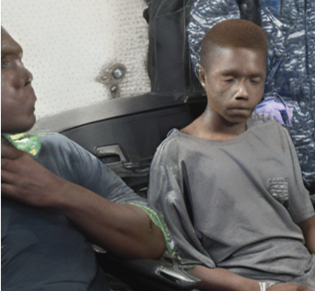} &
					\includegraphics[width=0.18\textwidth]{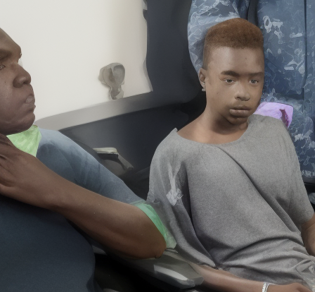} &
					\includegraphics[width=0.18\textwidth]{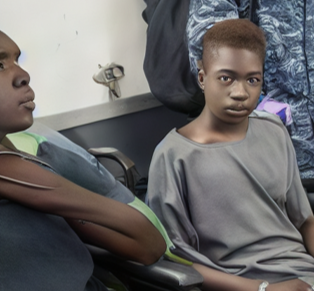} \\
					StableSR\cite{wang2023exploiting} &
					DiffBIR\cite{lin2023diffbir} &
					PASD\cite{yang2023pixel} \\
					\includegraphics[width=0.18\textwidth]{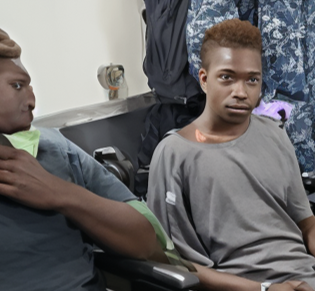} &
					\includegraphics[width=0.18\textwidth]{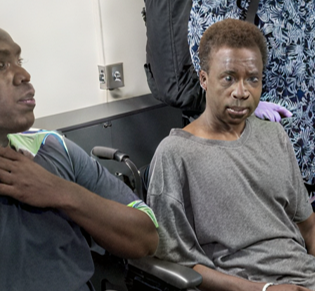} &
					\includegraphics[width=0.18\textwidth]{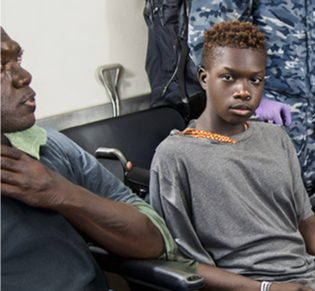} \\
					SeeSR &
					SUPIR &
					\method{EAM (ours)}
				\end{tabular}
			\end{adjustbox}
			\hspace{-2mm}
		\end{tabular}
	}
	\vspace{-4mm}
	\caption{Qualitative comparison with different methods on DIV2K. Our EAM can restore the texture and details. Other methods recover with unpleasing details or irregular faces.}
	\label{fig:main_visual}
	\vspace{-4mm}
\end{figure}

\begin{figure}
	\centering
	\includegraphics[width=0.8\textwidth]{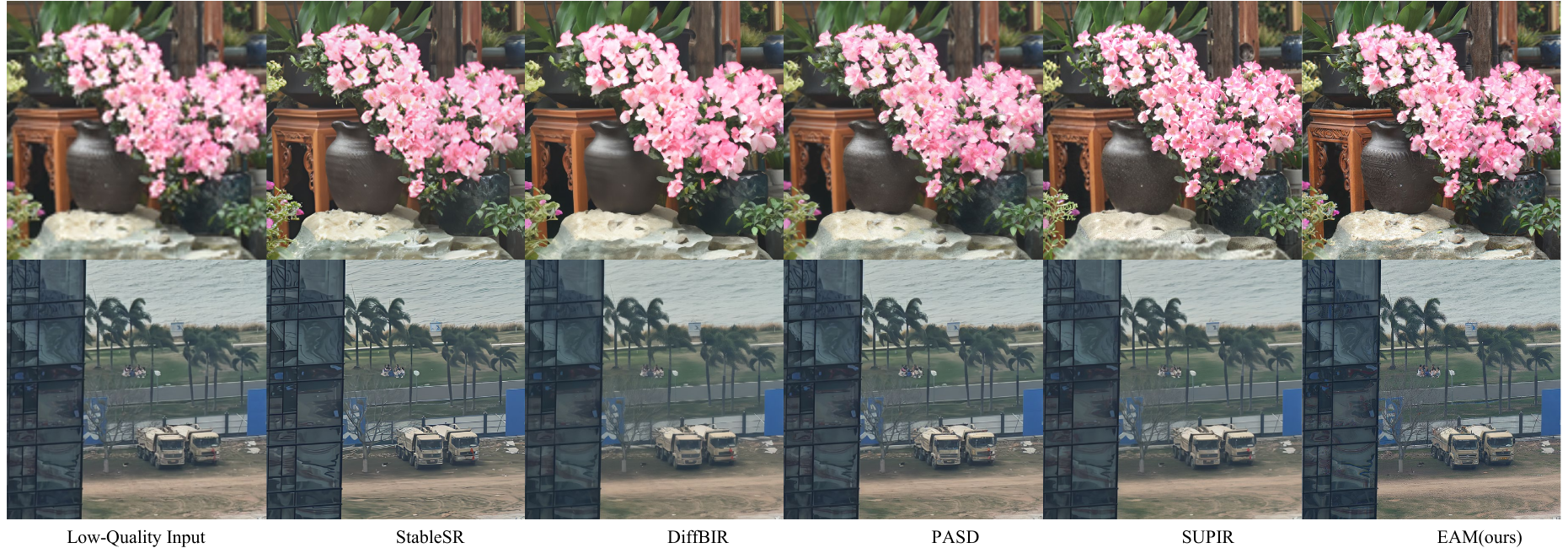}
	\caption{Qualitative comparison with different methods on real world inputs. Our EAM can restore the inputs with more realistic details.}
	\label{fig:RealWorld Compare}
\end{figure}

\begin{figure}
	\centering
	\includegraphics[width=0.6\textwidth]{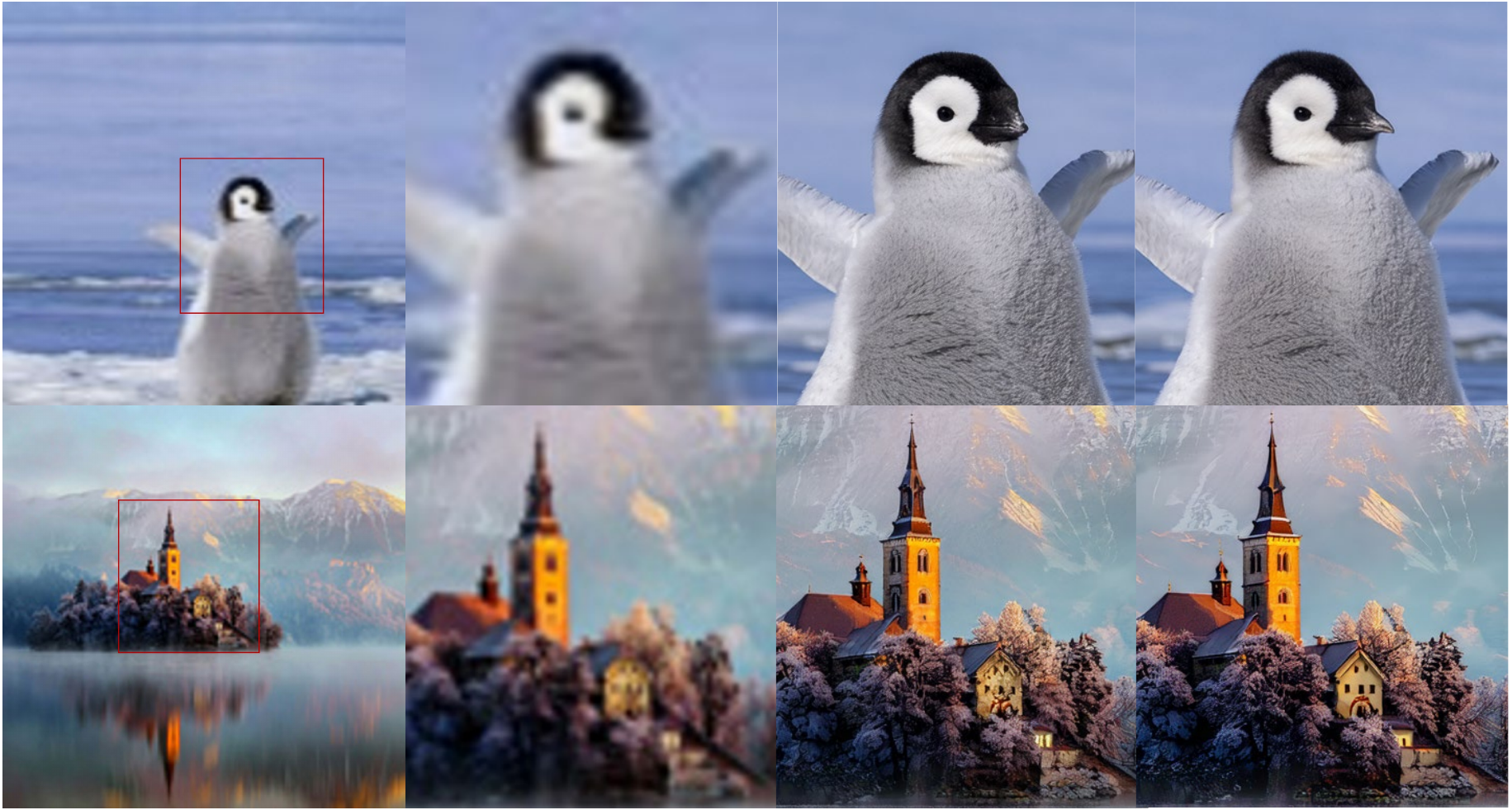}
	\caption{Restoration results using different architecture, from the left to the right: the whole LR input, the cropping input region, the corresponding restoration of DiT-controlnet, the result of EAM (Ours).}
	\label{fig:ab1}
\end{figure}

In comparison to the current state-of-the-art models such as SUPIR, our model achieves a more favorable balance between fidelity and generativity. The EAM structure, equipped with multiple advanced techniques, excels in regions with clear semantics, such as hair, vegetation, and architectural structures.

As shown in the accompanying Fig. \ref{fig:main_visual}, our model demonstrates superior realism and clarity in animal fur, surpassing other methods. It also exhibits good generative capabilities on medium to small faces without any facial distortion or noticeable facial defects.

As depicted in Fig.~\ref{fig:RealWorld Compare}, leveraging the powerful prior capabilities of SD3, our model shows significant competitiveness in terms of scene authenticity and detail. Moreover, it possesses an extremely strong generalization ability, truly achieving super-resolution for all things. Fig. \ref{fig:ab1} further corroborate the model's advanced super-resolution capabilities, with no apparent weaknesses in its performance.

\subsection{Ablation Study}
\label{exp_abl}
\paragraph{Comparison with ControlNet.}

\begin{table}[t]
	\centering
	\begin{tabular}{@{}lcccccccc@{}}
		\toprule
		\textbf{Architecture} & \textbf{Best Iters} & \textbf{PSNR} & \textbf{SSIM} & \textbf{LPIPS} & \textbf{ManIQA} & \textbf{ClipIQA} & \textbf{MUSIQ} \\ \midrule
		DiT-ControlNet        & 80k                & 20.07         & 0.5291        & 0.318          & 0.6502         & 0.703          & 73.25         \\
		$\Psi$-DiT            & 30k                & 19.75         & 0.5197        & 0.3329         & 0.6584         & 0.7142         & 73.86         \\ 
		\bottomrule
	\end{tabular}
	\caption{Ablation on the key architecture.}
	\label{tab:psi}
\end{table}

\begin{table}[h]
	\centering
	\begin{tabular}{@{}ccc|ccccccc@{}}
		\toprule
		\textbf{Zero Init} & \textbf{TEB Init} & \textbf{NLB Init} & \textbf{PSNR} & \textbf{SSIM} & \textbf{LPIPS} & \textbf{ManIQA} & \textbf{ClipIQA} & \textbf{MUSIQ} \\ \midrule
		$\times$           & $\times$          & $\times$          & 19.66         & 0.5089        & 0.3669         & 0.6331         & 0.6816         & 73.18         \\
		\checkmark         & $\times$          & $\times$          & 19.61         & 0.5117        & 0.342          & 0.6362         & 0.6899         & 73.24         \\
		\checkmark         & \checkmark        & $\times$          & 19.87         & 0.5183        & 0.3276         & 0.6511         & 0.7023         & 73.39         \\
		\checkmark         & $\times$          & \checkmark        & 19.75         & 0.5197        & 0.3329         & 0.6584         & 0.7142         & 73.86         \\ 
		\bottomrule
	\end{tabular}
	\caption{Methods for Initializing the Conditional Branch, where TEB Init refers to the initialization from the text embedding branch, and NLB Init indicates the initialization from the noisy latent branch.}
	\label{tab:init}
\end{table}

\begin{table}[h]
	\centering
	\begin{tabular}{@{}lcccc@{}}
		\toprule
		Mask Strategy  & ManIQA & ClipIQA & MUSIQ \\ \midrule
		No Mask        & 0.6584 & 0.6813 & 73.14 \\
		Fix 0.50       & 0.7051 & 0.6826 & 73.08 \\
		Fix 0.75       & 0.6897 & 0.6916 & 73.20 \\
		Fix 0.85       & 0.7048 & 0.6916 & 72.32 \\
		Random 0.75–1.00 & 0.7072 & 0.6953 & \textbf{73.60} \\
		PMS (ours)     & \textbf{0.7077} & \textbf{0.6984} & 73.44 \\
		\bottomrule
	\end{tabular}
	\caption{Comparison of different MIM strategies on NTIRE2024-RAM50. Best performance is highlighted in bold.}
	\label{tab:pms_comp}
\end{table}

\begin{table}
	\centering
	\begin{tabular}{@{}ccccc@{}}
		\toprule
		Training & Testing  & ManIQA & ClipIQA & MUSIQ \\
		\midrule
		DA & DA &  0.6326 & 0.7333 & 73.13 \\
		DA & SA & 0.6301   & 0.7326 & 72.69  \\
		SA & SA & \textbf{0.6416} & \textbf{0.7652} & \textbf{73.95} \\
		\bottomrule
	\end{tabular}
	\caption{Comparison on different prompts training \& testing strategy on RealPhoto60. \textbf{Bold} represents the best performance. DA refers to detail-aware prompts, which is the traditional prompts generation method. SA refers to subject-aware prompts, which applied our new pipeline of prompts generation.}
	\label{tab:prompts_comp}
\end{table}

Despite the fact that the DiT architecture augmented with a ControlNet branch~\cite{zhang2023addingconditionalcontroltexttoimage} yields commendable results after 80k iterations, our proposed $\Psi$-DiT achieves convergence with only 30k iterations. By leveraging the MAE training paradigm, $\Psi$-DiT not only conserves training costs but also enhances convergence speed. The token-based conditional input branch offers greater flexibility and is more compatible with the MAE training approach.
As shown in Tab.~\ref{tab:psi}, on the DIV2K dataset, $\Psi$-DiT surpasses DiT-ControlNet in terms of ClipIQA and MUSIQ metrics. On the open-source dataset NTIRE2024 RAM50, $\Psi$-DiT leads across ManIQA, ClipIQA, and MUSIQ metrics, thereby demonstrating its robust generative capabilities and excellent generalization in diverse scenarios.
Furthermore, as indicated in Tab.~\ref{tab:init}, our zero-initialization strategy ensures the maximization of the inherent generative potential of the original SD3, preserving the information contained in the parameters of the original DiT structure from the outset of training. We have meticulously selected the parameters for the noisy latent branch (NLB) to initialize the conditional branch. Compared to initialization with a text embedding branch (TEB), NLB initialization better adapts the conditional branch to LR image tokens.

\paragraph{Progressive MIM Strategy.}
To better integrate Masked Image Modeling (MIM) into our model, we conducted experiments with various mask ratios to comprehensively assess their influence on model performance. As illustrated in Tab.~\ref{tab:pms_comp}, comparing fixed mask ratios of 0.5 and 0.75, we observe that employing a mask ratio that is too low diminishes the model's generative capacity, consequently leading to a decline in metrics. We then explored several approaches to increasing the mask ratio. One method involved maintaining a high mask ratio of 0.85, another utilized a random mask ratio ranging from 0.75 to 1.0, and the third employed our proposed Progressive Masked Image Modeling (PMS) strategy. Our analysis revealed that, compared to the default mask ratio of 0.75, PMS not only achieved superior performance metrics but also reduced the training costs for the network. This result further substantiates the advantages of PMS in transitioning from T2I models to I2I models.

\paragraph{Subject-aware Prompts.}
The efficacy of subject-aware prompts is shown in Tab.~\ref{tab:prompts_comp}. The table indicates that if the model inference relies solely on prompts collected by the subject-aware method, its performance is sub-optimal. However, by incorporating these subject-aware prompts during the training phase, the model can fully leverage its subject-aware capabilities, thereby enhancing the final result.

\section{Conclusion}
\label{sec:Conclusion}
With the advent of the Diffusion Image Transformer (DiT) as a formidable diffusion network, the emergence of DiT-based super-resolution (SR) models is imminent. It is anticipated that within these DiT-based SR networks, the constraints of the ControlNet will emerge as a bottleneck, hindering the achievement of exceptional SR performance. To counteract this limitation, we introduce the Enhancing Anything Model (EAM), a novel DiT-based architecture designed for blind image super-resolution.
Leveraging a pre-trained text-to-image DiT, we construct a triple-flow DiT block, denoted as $\Psi$-DiT, which integrates the noise latent variable with the text embedding and low-resolution (LR) latent variable to enhance attention mechanisms. Our ablation studies demonstrate that these novel blocks outperform the conventional ControlNet. Furthermore, during training, we incorporate the Progressive MIM Strategy to enhance the generative capacity and robustness of our model. Additionally, we employ a new subject-aware prompt generation pipeline to supply more insights to the model through in-context learning, which significantly benefits the SR results.
Extensive experiments confirm that our model excels over current SR methodologies on real-world datasets.


\end{document}